\documentclass[letterpaper]{article}
\usepackage{iccc}
\usepackage{listings}
\usepackage{url}
\usepackage{todonotes}
\usepackage{courier}
\usepackage{pgf, tikz}
\usepackage{adjustbox}
\usepackage{booktabs}
\usepackage{xspace}
\usetikzlibrary{arrows, automata}
\usepackage{algorithm}
\usepackage{algorithmic}

\newcommand{\gi}{\textsc{Gitta}\xspace}

\usepackage{times}
\usepackage{helvet}
\usepackage{courier}
\title{Discovering Textual Structures: \\Generative Grammar Induction using Template Trees}
\author{Thomas Winters \and Luc De Raedt \\
Computer Science Department, Leuven AI\\
KU Leuven, Belgium\\
\{firstname\}.\{lastname\}@kuleuven.be\\
}
\begin{document} 
\maketitle

\begin{abstract}
\begin{quote}

 Natural language generation provides designers with methods for automatically generating text, e.g. for creating summaries, chatbots and game content.
 In practise, text generators are often either learned and hard to interpret, or created by hand using techniques such as grammars and templates. %
 In this paper, we introduce a novel grammar induction algorithm for learning interpretable grammars for generative purposes, called \gi.
 We also introduce the novel notion of template trees to discover latent templates in corpora to derive these generative grammars. %
 By using existing human-created grammars, we found that the algorithm can reasonably approximate these grammars using only a few examples.
 These results indicate that \gi could be used to automatically learn interpretable and easily modifiable grammars, %
 and thus provide a stepping stone for human-machine co-creation of generative models.

\end{quote}
\end{abstract}

\section{Introduction}

Text generation is a prominent tool within computational creativity, due to many creative fields using text as its primary medium, e.g. poetry and humor.
As such, many computational creative systems have applied a large variety of text generation methods.
Methods like templates and grammars generally grant the model designer relatively more control over the output, but tend to be labor-intensive to create.
Neural text generation approaches (e.g. RNNs, transformer models) on the other hand can create impressive language generators, but at the cost of predictability, interpretability and ease of regulating its outputs in a directed way.
In this paper, we explore a new technique for learning grammars for generative purposes by discovering latent templates such that the grammar is easily interpretable and modifiable by designers of generative textual models for creative purposes.

\section{Background}

\subsection{Templates}

Using templates is a popular approach for generating text.
In this context, a template is a piece of text with several slots that are later filled in using a particular data source.
While they have a lot of obvious merits for e.g. chatbots and front-end web development, templates are also popular in computational creative applications.
For these creative purposes, templates are often paired with schemas for providing sensible content to the templates that internally encode how the template slot values relate to each other \cite{binsted1994jape,winters2018automaticjokegeneration}.
For example, the expansion of non-terminal $S$ in Figure~\ref{fig:grammar-example} could be seen as a template, where a schema \emph{(not in the figure)} would create sensible pairings for $T$ and $F$.
There have been several efforts into automatically learning such templates and schemas from single examples by analysing 
linguistic relationships and properties \cite{hong2009tpeg,winters2019torfsbot}.

\subsection{Generative Grammars}

Grammars are another popular way of generating text.
A context-free grammar (CFG) is a four-element tuple $(V, \Sigma, R, S)$, where $V$ is a finite set of non-terminals, $\Sigma$ a finite set of terminals, $R$ a set of production rules that map elements of $V$ to $(V \cup \Sigma)^{*}$ and $S$ the start symbol.
While generative grammars were initially mainly used for generating according to a text need, they are also used for creative purposes.
Tracery is a popular language among casual creators for designing generative grammars.
Such grammars usually extend CFGs, e.g. adding stored assignments and rule weights \cite{compton2015tracery,winters2019samsonbot}.
A prominent design pattern in these grammars is specifying production rules that map non-terminals either to templates, or to a list of possible values for a particular template slot (Figure \ref{fig:grammar-example}).
The grammar then fills the templates with randomly generated slot value combinations.

\begin{figure}[h!]
  \centering
  \begin{lstlisting}
    S -> I like putting <T> on my <F>
    T -> cheese | pineapple | soy sauce
    F -> pizza | salad | muesli | sushi
  \end{lstlisting}
  \caption{An example grammar capable of generating 12 different sentences specifying (odd) dish toppings.}
  \label{fig:grammar-example}
\end{figure}

\subsection{Learning Grammars}

There are many different algorithms for inducing CFGs, usually designed for a particular class of grammar.
The most popular type of grammar induction induces part-of-speech tag structures from treebanks or plain text. %
Another popular type of grammar induction is discovering repetitive structures to help encode the input text efficiently, inducing grammars where each non-terminal has only a single production rule \cite{nevill1997sequitor}.
The grammars induced by these algorithms are however
shaped differently than typical generative grammars with the template-like production rules.
The latter generally avoids recursive production rules, as generating texts of unbounded lengths is usually undesirable for the creative goal. %
Non-recursive grammars are thus tools for compactly specifying a finite space or interesting texts. %
In this paper, we introduce an algorithm that can learn such non-recursive context-free grammars using a template-focused approach, which can thus easily be interpreted and adapted by generative grammar creators.

\section{Template Trees}

We create and define the notion of template trees as an intermediary step for inducing a generative grammar, and propose an algorithm for learning template trees from input text.

\subsection{Template Tree Definition}

\begin{figure}[h]
    \centering
    
\begin{tikzpicture}[
            > = stealth, %
            shorten > = 1pt, %
            auto,
            node distance = 2cm, %
            semithick %
        ]

        \node (r) [] {$\langle A \rangle$};
        \node (g1) [below left=0.2cm and 1.7cm of r] {hello $\langle B \rangle$};
        \node (g2) [right of=g1] {$\langle C \rangle$ world};
        \node (g3) [right of=g2] {hi $\langle D \rangle$};
        \node (g4) [right of=g3] {$\langle E \rangle$ people};

        \node (s1) [below=0.4cm and 1.7cm of g1] {hello world};
        \node (s2) [right of=s1] {hello people};
        \node (s3) [right of=s2] {hi world};
        \node (s4) [right of=s3] {hi people};

        \path[->] (r) edge node {} (g1);
        \path[->] (r) edge node {} (g2);
        \path[->] (r) edge node {} (g3);
        \path[->] (r) edge node {} (g4);
        
        \path[->] (g1) edge node {} (s1);
        \path[->] (g1) edge node {} (s2);
        \path[->] (g2) edge node {} (s1);
        \path[->] (g2) edge node {} (s3);
        \path[->] (g3) edge node {} (s3);
        \path[->] (g3) edge node {} (s4);
        \path[->] (g4) edge node {} (s2);
        \path[->] (g4) edge node {} (s4);
 
\end{tikzpicture}

    \caption{A template tree example}
    \label{fig:tt-example}
\end{figure}
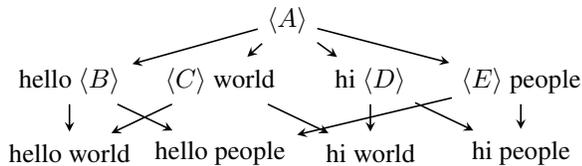

A template tree is a connected acyclic directed graph where each node represents a template that is more general than the template of all its child nodes, thus defining a partial ordering.
The leaves of a template tree are templates without slots, i.e. the input sentences used to learn this tree.
A template slot maps to zero or more other template elements (i.e. slots and/or word tokens).
A simple template tree can be seen in Figure \ref{fig:tt-example}.

\subsection{Learning Template Trees}

\subsubsection{Merging Templates}
To create a template for the parent node of two templates nodes, we use a longest common subsequence algorithm on their word tokens to create the most specific template that is more general than its children. %
More specifically, we adapted the Wagner-Fischer algorithm and use the displacement matrix to insert slots when tokens differ \cite{wagner1974string}.
When there are multiple longest subsequences, the algorithm ignores templates that are longer than the original templates\footnote{E.g., when merging "$hi\: \langle X \rangle$" and "$\langle Y \rangle \: hi$", it discards "$\langle Y \rangle \: hi \: \langle X \rangle$"
to avoid overgeneralisation, and prefers "$\langle Z \rangle$".}, and prefers less slots.

\subsubsection{Template Distance}

The distance between two templates is defined in terms of the merged template $m_{ij}$ causing the lowest distance $d$ to templates $t_i$ and $t_j$:
$$ d(t_i, t_j) = max(l_{t_i}, l_{t_j} ) - l_{m_{ij}} + s_{m_{ij}} - min(s_{t_i}, s_{t_j}) $$
where $l_t$ is the number of non-slot elements of a template $t$, and $s_t$ the number of slots in $t$.

\subsubsection{Learning Algorithm}

Algorithm \ref{alg:tt-learning} shows how a template tree is learned from input texts $D$.
Initially, all pairs of input texts are stored in a priority queue $Q$, sorted by their distance as defined above.
The algorithm keeps track of active templates, i.e. templates without parent, in a list $A$.
As long as $A$ contains more than one template, the algorithm will take all minimally distant pairs of templates from $Q$, where both templates are still active.
Every such minimally distant pair is merged, making both templates inactive and the new merged template active.
All new templates are then paired up with other active templates, and added to $Q$.
If $A$ only contains one template, then this template becomes the root of the template tree.
The template tree is reconstructed by adding all templates that once merged to a particular template as children of the node of this template.

\begin{algorithm}[h]
\begin{algorithmic}
\REQUIRE input texts $D$
\ENSURE $A=\{t\}$ where $leaves_{t} = D$

\STATE $Q \leftarrow \{ (d_i, d_j) \mid  d_i,d_j \in D \}$
\STATE $A \leftarrow D$

\WHILE{$\#A > 1$}
\STATE $M \leftarrow argmin_{(t_i, t_j) \in Q}( d(t_i, t_j) )   \cap    \{ (a,b) \mid  a, b \in A \} $
\STATE $N \leftarrow \{\} $
    
\FORALL{$ (t_i, t_j) \in M$}
\STATE{$m \leftarrow merge(t_i, t_j) $}
\STATE{$N \leftarrow N \cup \{ m \}$}
\STATE{$A \leftarrow A \setminus \{t_i, t_j\}$}
\ENDFOR

\FORALL{$ n \in N$}
\STATE{$Q \leftarrow Q \cup \{ (n, a) \mid n \in N, a \in A \}$}
\STATE{$A \leftarrow A \cup \{ n \}$}
\ENDFOR

\ENDWHILE

\end{algorithmic}
\caption{Calculating template tree from input texts}
\label{alg:tt-learning}
\end{algorithm}

\section{\gi: Template Tree to Grammar}

We introduce a new grammar induction algorithm named \gi \emph{(Grammar Induction using a Template Tree Approach)}.
\gi aims to induce a non-recursive CFG, thus compactly representing a finite number of similar finite strings.  %
While any finite language of size $n$ can trivially be represented by a simple grammar with $n$ production rules, having fewer production rules implies that patterns have been induced.
This allows the grammar to potentially generate unseen examples from the language, and also be more easily modifiable.
\gi converts the template tree into a grammar by assuming independence between slot values, and simplifying the template tree. %
The resulting slot values and root template then specify the grammar. %

\subsection{Pruning Template Tree}

\gi first prunes redundant children of template tree nodes.
A child is redundant if all its descendant leaves are reachable through the other children.
For each level, nodes are checked in ascending order of number of descendants, pruning nodes with less general templates first.

\subsection{Merging Slots}

To convert the template tree into a grammar, \gi assumes that all possible slot values are independent from all other slot values of the template.
For every slot, all possible slot values are extracted from the templates of the children of the nodes having a template with this slot.

After finding all slot values $U_i$ for every slot $s_i$, the algorithm merges similar slots if $\frac{\#(U_i \cap U_j)}{\#(U_i \cup U_j)} \geq r$,
where $r \in [0,1]$ is determined by the user.
Lower values of $r$ thus require slots to have less overlap in slot values in order to be merged.
\gi also removes $u_j \in U_i$ if there is a slot $s_k$ such that $s_k \in U_i$ and $u_j \in U_k$.
If $s_i \in U_i$, the slot will also be removed from the slot values.
If $U_i = \{s_k\}$, then $s_i$ is replaced with $s_k$.
For example, for the tree in Figure \ref{fig:tt-example}, the algorithm would discover that $D$ has the same slot values as $B$, and thus should be replaced by $B$.
This process continues until there is an iteration without any update.

\subsection{Collapsing Template Tree}

Using the merged slots, several simplifications are made to the template tree.
First, the replacement mapping reduces the number of different slots of the template tree.
Second, knowing the slot values for a slot helps reduce the number of nodes in the tree.
For a node $p$ with template $t_p$, and a child $c$ with template $t_c$ that contains a slot of $t_p$ and for which the template $t_c$ can be obtained by filling in other slots with known slot values into $t_p$, then this child node $c$ is redundant and can be pruned.
All children of $c$ are then added as direct children of $p$.
For example, for Figure \ref{fig:tt-example}, if the root template would be ``$\langle C \rangle \: \langle B \rangle$'', the four middle nodes would collapse into their parent, leaving only ``$\langle C \rangle \: \langle B \rangle$'' as parent of the four leaf nodes.
After collapsing the template tree using knowledge of the slot values, the template of each node is recalculated, which leads to the aforementioned new root node template of Figure \ref{fig:tt-example}.
This process of simplification of the template tree and recalculation of the templates keeps repeating until the template tree is unchanged after an iteration.
The resulting grammar is derived by mapping from start symbol $S$ to the root template of the template tree, and using slot values mappings as production rules.

\section{Experiment: Reverse-engineering Grammars}

To measure the performance of the algorithm, we test how well it can induce grammars from generations of a human-made grammars.
We use Tracery grammars to generate a fixed number of sentences that serve as examples for \gi.
The algorithm also receives the depth of the original grammar as a parameter to limit the height of the template tree.
After inducing a grammar $I$, we compare how many elements that are generatable using $I$ are in the language $L_G$ defined by $G$, and how many elements of $L_I$ are not in $L_G$.
We also compare size of the grammars, i.e. number of production rules $\#R_I$ of the induced grammar to $\#R_G$ of the original grammar $G$.
These production rules can not have disjunctions on the right side, meaning a rule with the shape $A \rightarrow cat \mid dog$ would be normalised to the two rules $A \rightarrow cat$ and $A \rightarrow dog$.
Smaller grammars are generally more interpretable, and for \gi also an indication for how well the grammar compacted information.

Out of all Tracery bots on listed BotWiki\footnote{\url{https://botwiki.org/}} \emph{(159)}, we downloaded all sources that were available on CheapBotsDoneQuick\footnote{\url{https://cheapbotsdonequick.com/}} \emph{(58)} and used all grammars without advanced syntax \emph{(47)} that only generated text \emph{(31)} with at most one million possible generations \emph{(10)} in order to make it feasible to calculate  $L_G \cap L_I$.
We also removed non-terminal modifiers, used e.g. for capitalisation and pluralisation, from the grammars, leaving only the bare non-terminals.
We ran \gi five times on every grammar on randomized subsets of $L_G$ of size 25, 50 and 100 examples, and took the median values over the runs.

\begin{table*}[ht!]
\centering
\begin{tabular}{@{}llll|lll|lll|lll@{}}
\toprule
\multicolumn{4}{c|}{\textbf{Grammar $G$}}     & \multicolumn{3}{c|}{\textbf{$I$ from 25 examples}}   & \multicolumn{3}{c|}{\textbf{$I$ from 50 examples}} & \multicolumn{3}{c}{\textbf{$I$ from 100 examples}}                        \\ \midrule
\emph{id} & \emph{Name}   & \emph{$\# L_G$} & \emph{$\#R_G$} & \emph{$\in L_G$} & \emph{$\notin L_G$} & \emph{$\#R_I$} & \emph{$\in L_G$} & \emph{$\notin L_G$} & \emph{$\#R_I$} & \emph{$\in L_G$} & \emph{$\notin L_G$} & \emph{$\#R_I$} \\ \midrule
1 & botdoesnot      & 380292                  & 363           & 648              & 0                    & 64            & 2420             & 0                    & 115           & 1596             & 4                    & 179           \\
2 & BotSpill        & 43452                   & 249           & 75               & 0                    & 32            & 150              & 0                    & 62            & 324              & 0                    & 126           \\
3 & coldteabot      & 448                     & 24            & 39               & 0                    & 38            & 149              & 19                   & 63            & 388              & 9                    & 78            \\
4 & hometapingkills & 4080                    & 138           & 440              & 0                    & 48            & 1184             & 3240                 & 76            & 2536             & 7481                 & 106           \\
5 & InstallingJava  & 390096                  & 95            & 437              & 230                  & 72            & 2019             & 1910                 & 146           & 1156             & 3399                 & 228           \\
6 & pumpkinspiceit  & 6781                    & 6885          & 25               & 0                    & 26            & 50               & 0                    & 54            & 100              & 8                    & 110           \\
7 & SkoolDetention  & 224                     & 35            & 132              & 0                    & 31            & 210              & 29                   & 41            & 224              & 29                   & 49            \\
8 & soundesignquery & 15360                   & 168           & 256              & 179                  & 52            & 76               & 2                    & 83            & 217              & 94                   & 152           \\
9 & whatkilledme    & 4192                    & 132           & 418              & 0                    & 45            & 1178             & 0                    & 74            & 2646             & 0                    & 108           \\
10& Whinge\_Bot     & 450805                  & 870           & 3092             & 6                    & 80            & 16300            & 748                  & 131           & 59210            & 1710                 & 222           \\ \bottomrule
\end{tabular}

\centering
\caption{Grammar induction results given a specific number of random generations of $G$, measuring median number of generations of the induced grammar $I$ that are in and not in the target language, as well as their median sizes, over five runs.}
\label{tab:increasing-induction}
\end{table*}

As can be seen in Table \ref{tab:increasing-induction}, the algorithm is generally able to induce grammars that generate significantly more elements of the original language than shown as example to the algorithm, with usually relatively few elements not in the original language.
However, \gi also sometimes uses relatively more rules $R_I$ to generate relatively less generations $L_I$, most notably in grammars 1 and 2.
This indicates that many rules are likely redundant or should be decomposed into simpler rules to allow for more generations.
For grammar 6, generalisation is not possible due to the origin template having one slot, and this slot mapping to different word lists, which also explains why it has more production rules than generations.

For grammars 4 and 5, \gi tends to induce grammars with relatively large numbers of generations that are not in $L_G$.
This is usually due to overgeneralisation.
For example, a grammar $G$ that has the production rule ``$S \rightarrow \langle Hello \rangle \; \langle World \rangle \mid \langle Hello \rangle \; there, \: \langle Name \rangle$'', might lead to \gi creating a more general rule ``$S \rightarrow \langle Hello \rangle \: \langle There \rangle \: \langle Thing \rangle$'', with ``$There \rightarrow there \mid \epsilon$'' and $ Thing $ mapping to all values of $Name$ and $World$.
For grammar 5 in particular, the origin template has four consecutive non-terminals separated from two other non-terminals by only one terminal, all mapping to varying number of terminals.
This property makes it unclear for \gi where slots start and end, thus leading to overly specific production rules being added instead of finding clear slot values.

\section{Discussion \& Future Work}

\gi could be employed in a collaborative generative grammar building tool, where a designer and the algorithm create a generative grammar together.
In this scenario, the designer could first illustrate several examples or use an existing corpus specifying what the grammar should generate, for which the algorithm will propose a suitable grammar by discovering latent templates, thus creating an initial grammar prototype.
The designer can then add, remove and modify production rules to further suit their needs, thus allowing more meaningful interactions than black-box generative text generators generally allow.
This direct control could be used e.g. for limiting the possibilities of generating offensive or unwanted content, which is an important aspect for many text generation domains such as game development.

One limitation compared to other grammar induction algorithms is that it cannot induce recursive grammars.
As such, production rules like $S \rightarrow SS \mid (S) \mid \epsilon$ \emph{(= the bracket language)} are not able to be learned by our system.
However, since recursion is generally an unwanted property of generative grammars due to making grammars able to generate unbounded texts, our proposed algorithm thus prevents language model overgeneralization caused by recursion.

\gi creates a basis for learning more complex, interpretable generative models.
It could be trivially extended by learning probabilities of rules as a post-processing step using the input sentences.
Another interesting extension is learning constraints that hold between expansions of non-terminals, and thus create complex generative schemas. %

We mainly see the use for this algorithm in automatically mimicking patterns or extending data sets that have some sort \emph{(possibly latent)} template in their texts, such as forum topic titles or writing and comedy prompts.
Template trees in itself could also be used for discovering frequently occurring templates in a corpus, and provide similar functionality as clustering algorithms.
The code of \gi is available on \url{https://github.com/twinters/gitta}. %

\section{Conclusion}

We introduced a new way for learning context-free grammars, focusing on interpretability and its generative performance.
We introduced the notion of template trees to achieve this purpose, as well as a learning algorithm for this structure and transformations.
The experiments indicate that the grammar induction algorithm is able to induce real grammars from little examples, showing its potential for use in collaborative modelling of grammars.
We hope that this system could be a stepping stone towards automatic co-creation of complex but interpretable generative grammars. %

\section{Acknowledgments}
Thomas Winters is a fellow of the Research Foundation-Flanders (FWO-Vlaanderen).

\bibliographystyle{iccc}
\bibliography{references}

\end{document}